\documentclass[a4paper,fleqn]{cas-dc}

\usepackage[numbers]{natbib}

\def\tsc#1{\csdef{#1}{\textsc{\lowercase{#1}}\xspace}}
\tsc{WGM}
\tsc{QE}
\tsc{EP}
\tsc{PMS}
\tsc{BEC}
\tsc{DE}

\usepackage{amssymb}
\usepackage{amsmath}
\usepackage{subcaption}
\usepackage{multirow}
\usepackage{mathrsfs}
\usepackage{silence}
\usepackage{float}
\usepackage{hyperref}
\usepackage[hypcap=false]{caption}
\usepackage{placeins}
\usepackage{etoolbox}
\makeatletter
\patchcmd{\bio}{\includegraphics}{\includegraphics[width=2.5cm,height=3cm,keepaspectratio=false]}{}{}
\makeatother

\begin{document}
\let\WriteBookmarks\relax
\def\floatpagepagefraction{1}
\def\textpagefraction{.001}

\shorttitle{MNAS-Unet for medical image segmentation}
\shortauthors{L. Meng et~al.}

\title[mode=title]{Optimizing Neural Network Architecture for Medical Image Segmentation Using Monte Carlo Tree Search}

\author[1,3]{Liping Meng}
\cormark[1]
\ead{lipingmeng1999@gmail.com}
\orcidauthor{0009-0001-3631-1017}{Liping Meng}

\affiliation[1]{organization={Xi'an Kedagaoxin University, Faculty of Science and Engineering},
                city={Xi'an},
                postcode={710109},
                state={Shaanxi},
                country={China}}

\author[2]{Fan Nie}
\author[2]{Yunyun Zhang}
\author[1]{Chao Han}

\affiliation[2]{organization={Xi'an Kedagaoxin University, School of International Education and Humanities},
                city={Xi'an},
                postcode={710109},
                state={Shaanxi},
                country={China}}

\affiliation[3]{organization={Saint Petersburg State University},
                city={Saint Petersburg},
                postcode={199034},
                country={Russia}}

\cortext[cor1]{Corresponding author}

\begin{abstract}
This paper proposes a novel medical image segmentation framework, MNAS-Unet, which combines Monte Carlo Tree Search (MCTS) and Neural Architecture Search (NAS). MNAS-Unet dynamically explores promising network architectures through MCTS, significantly enhancing the efficiency and accuracy of architecture search. It also optimizes the DownSC and UpSC unit structures, enabling fast and precise model adjustments. Experimental results demonstrate that MNAS-Unet outperforms NAS-Unet and other state-of-the-art models in segmentation accuracy on several medical image datasets, including PROMISE12, Ultrasound Nerve, and CHAOS. Furthermore, compared with NAS-Unet, MNAS-Unet reduces the architecture search budget by 54\% (early stopping at 139 epochs versus 300 epochs under the same search setting), while achieving a lightweight model with only 0.6M parameters and lower GPU memory consumption, which further improves its practical applicability. These results suggest that MNAS-Unet can improve search efficiency while maintaining competitive segmentation accuracy under practical resource constraints.
\end{abstract}

\begin{graphicalabstract}
\end{graphicalabstract}

\begin{highlights}
\item Novel MCTS-based architecture search framework for medical image segmentation
\item Achieved superior performance on multiple medical datasets (PROMISE12, CHAOS, Ultrasound)
\item Reduced architecture search cost by $\sim$54\% (139 vs.\ 300 search epochs) compared to NAS-Unet
\item Lightweight model with only 0.6M parameters and lower GPU memory consumption
\item Resource-efficient architecture suitable for deployment under limited GPU memory
\end{highlights}

\begin{keywords}
Monte Carlo Tree Search \sep Neural Architecture Search \sep Medical image segmentation \sep U-Net \sep Deep learning \sep Computer-aided diagnosis
\end{keywords}

\maketitle

\section{Introduction}
\label{sec1}

Medical image segmentation has become an indispensable tool in modern clinical practice, playing a crucial role in disease diagnosis, treatment planning, and patient monitoring. The accurate delineation of anatomical structures and pathological regions from medical images directly influences clinical decision-making and patient outcomes. However, manual segmentation of medical images is labor-intensive, time-consuming, and subject to inter-observer variability, making it impractical for large-scale clinical applications \cite{sun2023public,galbusera2024image}.

The complexity of medical images presents significant challenges for automated segmentation systems. Variations in imaging protocols, patient positioning, pathological conditions, and anatomical structures across different individuals create a highly heterogeneous dataset that requires robust and adaptive segmentation algorithms. Furthermore, the critical nature of medical applications demands high accuracy and reliability, as segmentation errors can lead to misdiagnosis or inappropriate treatment decisions.

To address these challenges, researchers have developed various deep learning approaches for medical image segmentation. U-Net, proposed by Ronneberger et al.\ (2015), has emerged as a fundamental architecture due to its encoder-decoder structure with skip connections, making it particularly suitable for medical imaging tasks \cite{ronneberger2015u}. However, manually designing optimal network architectures for specific medical imaging tasks remains a challenging and time-consuming process that typically requires extensive domain expertise and computational resources.

Building upon U-Net, researchers have proposed numerous variants to improve segmentation performance. Ange Lou (2021) introduced DC-UNet, which achieved more efficient feature extraction through dual-channel convolution structures \cite{lou2021dc}. Zhangfu Dong et al.\ (2022) presented MNet, combining 2D and 3D convolutions for multidimensional information fusion \cite{dong2022mnet}. The following year, Qing Xu et al.\ (2023) proposed DCSAU-Net, incorporating split-attention mechanisms to capture fine anatomical structures \cite{xu2023dcsau}, while the nnUNet team introduced MedNeXt (2023), leveraging large kernel architectures for enhanced global feature extraction \cite{roy2023mednext}. Most recently, Asmaa Sabet Anwar et al.\ (2025) proposed ResTransUNet, combining CNN and Transformer architectures for liver and tumor segmentation in CT images \cite{anwar2025restransunet}.

Despite these advances, existing U-Net variants face several limitations: (1) they require extensive manual design and hyperparameter tuning, (2) their architectures are often suboptimal for specific medical imaging tasks, and (3) they lack adaptability to different medical imaging modalities and anatomical structures.

To overcome these limitations, researchers have explored Neural Architecture Search (NAS) for automatically discovering optimal network architectures. NAS-Unet, proposed in 2019, explored differentiable NAS strategies (e.g., DARTS/ProxylessNAS-style methods) on a U-shaped backbone for medical image segmentation \cite{weng2019unet}. However, existing NAS methods face several challenges: (1) high computational requirements that limit practical applicability, (2) generic search spaces that do not account for medical imaging characteristics, and (3) insufficient consideration of clinical deployment constraints such as model size and inference speed.

Recent advances in NAS have attempted to address these challenges. Yang et al.\ (2020) proposed CARS, which used continuous evolution as an efficient alternative to differentiable NAS methods \cite{yang2020cars}. Yang et al.\ (2021) proposed DDSAS, which dynamically samples the search space and adapts the search process based on task requirements through a differentiable framework \cite{yang2021ddsas}. Miahi et al.\ (2022) explored genetic algorithms in neural architecture search, demonstrating their effectiveness in discovering efficient architectures for medical imaging tasks \cite{miahi2022genetic}. Yang et al.\ (2023) provided a comprehensive review of modern NAS techniques, highlighting the shift towards more efficient search strategies \cite{yang2023revisiting}. However, these methods still face limitations in terms of search efficiency and adaptability to different medical imaging modalities.

In this work, we propose MNAS-Unet, a novel framework that combines Monte Carlo Tree Search (MCTS) with Neural Architecture Search to address these limitations. Our approach introduces three key innovations: (1) a domain-specific search space designed for medical imaging tasks, (2) an efficient MCTS-based search strategy that balances exploration and exploitation \cite{swiechowski2023monte}, and (3) a lightweight architecture that maintains high segmentation accuracy while reducing computational requirements.

The main contributions of this work are as follows:
\begin{enumerate}
\item We integrate Monte Carlo Tree Search (MCTS) with Neural Architecture Search (NAS) for medical image segmentation, providing a more efficient search strategy.
\item We design a specialized search space with 6 Down POs, 4 Up POs, and 6 Normal POs, optimized for medical imaging characteristics.
\item We demonstrate improved search efficiency and segmentation accuracy on multiple medical image datasets.
\end{enumerate}

\FloatBarrier
\section{Related Work}
\label{sec2}

\subsection{Neural Architecture Search in Medical Image Segmentation}
Neural Architecture Search (NAS) has emerged as a promising approach for automatically discovering optimal network architectures in medical imaging applications. Traditional NAS methods, such as DARTS \cite{liu2019darts} and ProxylessNAS \cite{cai2019proxylessnas}, have shown success in general computer vision tasks but face significant challenges when applied to medical image segmentation. The high computational requirements and generic search spaces of these methods limit their practical applicability in medical imaging scenarios.

Recent advances in NAS have focused on adapting these methods specifically for medical applications. Swiechowski et al.\ (2023) provided a comprehensive review of Monte Carlo Tree Search (MCTS) applications in NAS, highlighting its potential for efficient architecture exploration in resource-constrained environments \cite{swiechowski2023monte}. MCTS provides an intelligent exploration strategy that balances exploitation of promising architectures with exploration of new search spaces, significantly reducing computational requirements while maintaining search quality.

\subsection{Medical Image Segmentation Architectures}
In medical image segmentation, U-Net \cite{ronneberger2015u} has become the de facto standard due to its encoder-decoder architecture with skip connections. Recent variants have focused on improving segmentation performance through various strategies. DC-UNet \cite{lou2021dc} introduced dual-channel convolution structures for more efficient feature extraction. MNet \cite{dong2022mnet} combined 2D and 3D convolutions for multidimensional information fusion, while DCSAU-Net \cite{xu2023dcsau} incorporated split-attention mechanisms to capture fine anatomical structures.

Recent work by Roy et al.\ (2023) introduced MedNeXt \cite{roy2023mednext}, leveraging large kernel architectures for enhanced global feature extraction. Anwar et al.\ (2025) proposed ResTransUNet \cite{anwar2025restransunet}, which combined CNNs and Transformers with pre-trained ResNet50 encoder for improved liver and tumor segmentation in CT images.

\subsection{Multi-scale Feature Fusion and Attention Mechanisms}
Multi-scale feature fusion and attention mechanisms have been widely applied in medical image segmentation to enhance model performance. Wang et al.\ (2024) proposed a multi-scale attention fusion network for retinal vessel segmentation, demonstrating the effectiveness of combining attention mechanisms with multi-scale feature extraction \cite{wang2024multi}. The network utilizes attention modules to focus on relevant anatomical structures while incorporating multi-scale feature fusion to capture both local and global context information.

However, these manually designed architectures often require extensive domain expertise and may not be optimal for specific medical imaging tasks. The integration of NAS with medical image segmentation presents an opportunity to automatically discover architectures tailored to the unique characteristics of medical images, including their high resolution, anatomical complexity, and clinical requirements for accuracy and interpretability.

\FloatBarrier
\section{Methodology}
\label{sec3}

\subsection{Neural Architecture Search}
Neural Architecture Search (NAS) \cite{yang2023revisiting} significantly reduces the manual cost of designing network structures and greatly expands the possibilities for network architecture design.

The algorithm for neural network architecture search typically encompasses three key components: the search space, the search strategy, and the strategy for performance evaluation \cite{miahi2022genetic}. As illustrated in Fig.~\ref{fig:neural_architecture_search}, the fundamental procedure of neural network architecture search involves the search strategy choosing a specific architecture $A$ from a set search space $\mathcal{A}$. This architecture is assessed by the performance evaluation module, which provides a performance estimate of $A$ back to the search strategy to inform the choice of the subsequent architecture. Ultimately, the algorithm produces a model $A^*$ that meets the preset performance requirements, serving as the output of the problem.

\begin{center}
\includegraphics[width=1.0\linewidth]{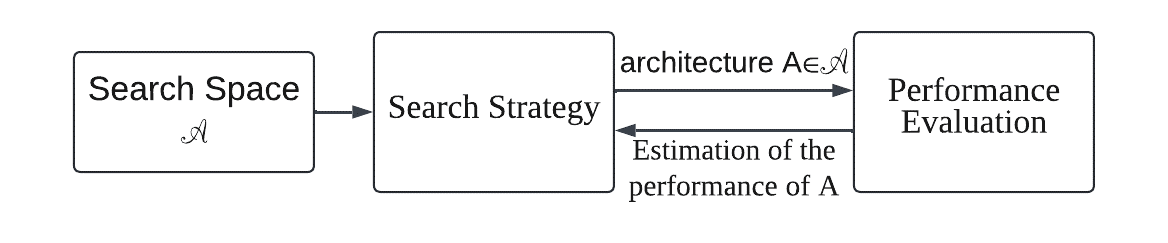}
\captionof{figure}{Neural architecture search.}
\label{fig:neural_architecture_search}
\end{center}

\subsection{U-Net Model}
The U-Net network \cite{ronneberger2015u}, proposed by Ronneberger et al.\ (2015), features a symmetric encoder-decoder architecture in a distinctive ``U'' shape. The encoder path extracts features and reduces spatial resolution through repeated $3 \times 3$ convolutions, ReLU activations, and $2 \times 2$ max pooling operations. This contracting path effectively doubles the feature map dimensions at each step while reducing spatial size, resulting in enhanced feature representation. The decoder path uses upsampling to gradually reconstruct spatial dimensions, accompanied by $3 \times 3$ convolutions and skip connections that concatenate features from corresponding encoder layers.

\subsection{Monte Carlo Tree Search (MCTS)}
The Monte Carlo Tree Search (MCTS) algorithm is a strategic search method that balances exploration and exploitation to effectively navigate large decision spaces \cite{wang2020neural,elkefi2024deep}. It involves four key steps: selection, expansion, simulation, and backpropagation. In the selection step, MCTS employs the Upper Confidence Bound for Trees (UCT) \cite{kocsis2006bandit} to determine the most promising nodes, ensuring a balance between exploring new possibilities and exploiting known successful paths. This iterative approach makes MCTS particularly effective for optimizing decision-making in complex domains \cite{kemmerling2024beyond}, such as neural architecture search and games, as shown in Fig.~\ref{fig:monte}.

\begin{center}
\includegraphics[width=0.8\linewidth]{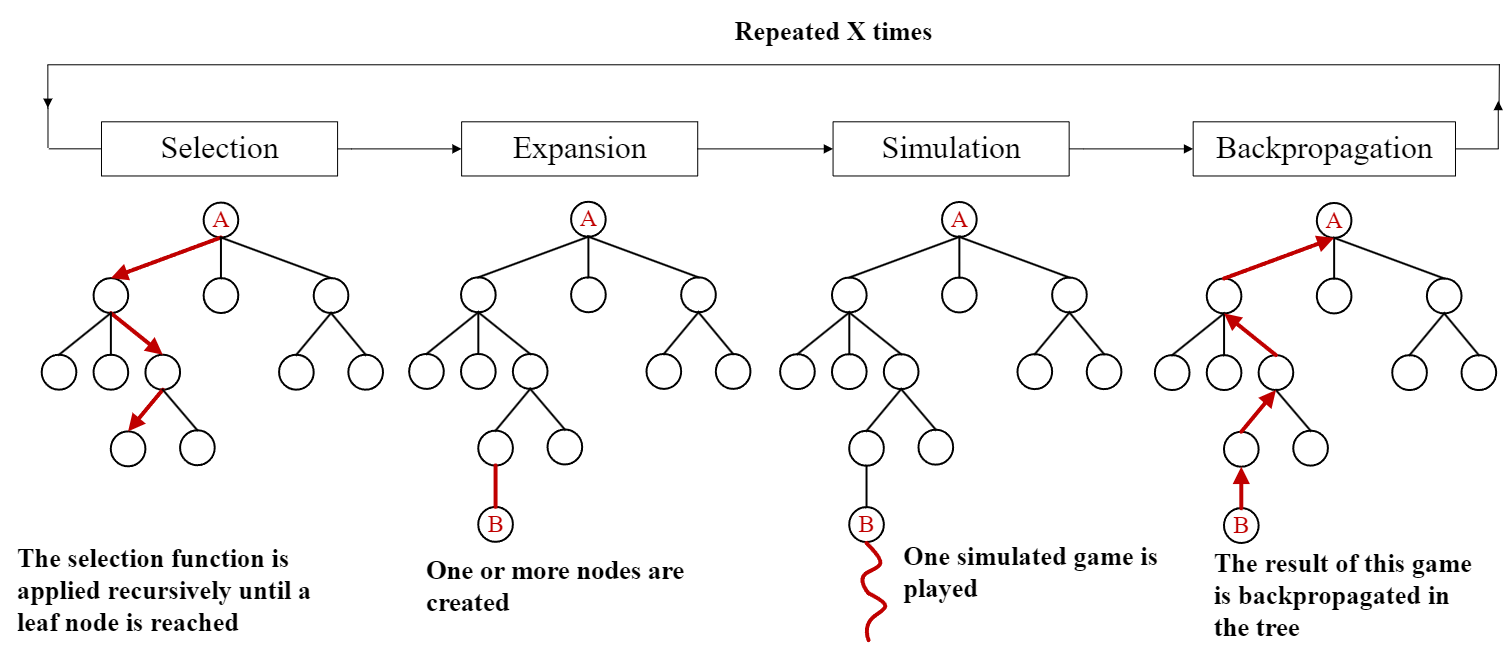}
\captionof{figure}{Monte Carlo Tree Search process (adapted from \cite{chaslot2008parallel}).}
\label{fig:monte}
\end{center}

\subsection{A Novel Network Architecture Search Model Based on MCTS: MNAS-Unet}

\subsubsection{Search Architecture of MNAS-Unet}
\begin{figure*}[tbp]
\centering
\vspace{0.8em}
\includegraphics[width=0.92\textwidth]{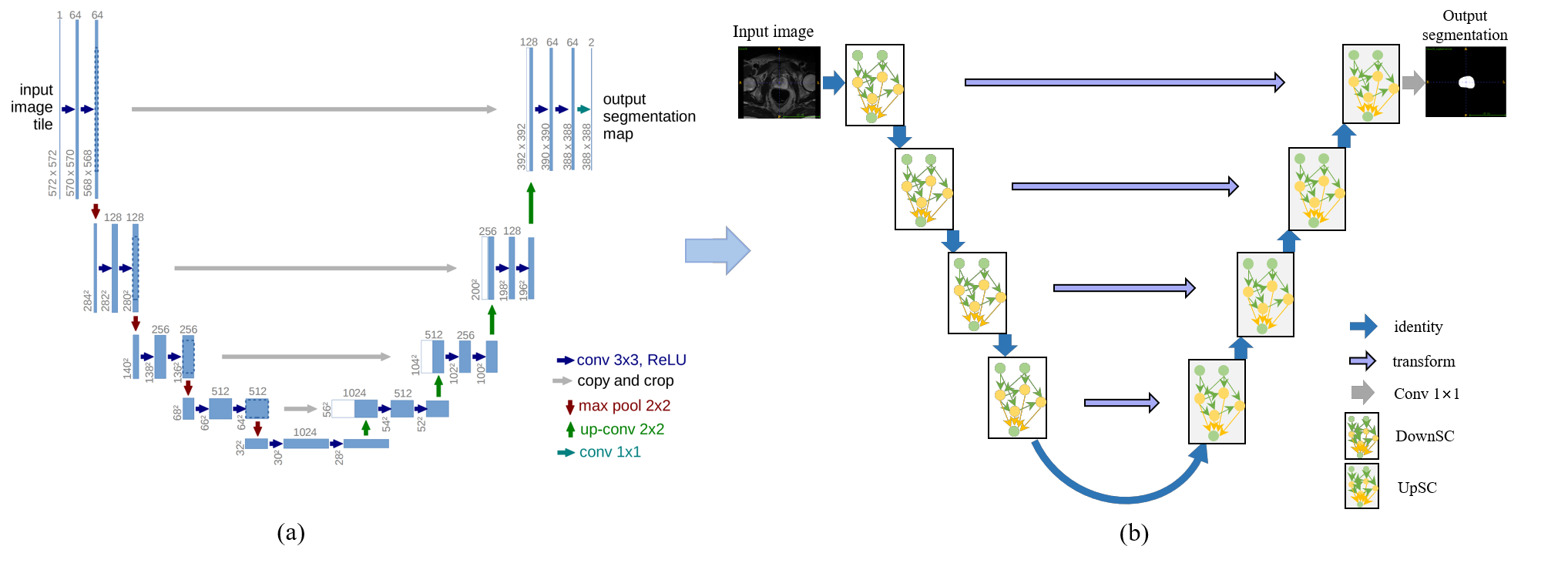}
\vspace{0.4em}
\caption{(a) Architecture of U-Net. (b) The U-shaped structure of the MNAS-Unet is depicted with rectangles, each representing a cell architecture subject to optimization. The specific operations of DownSC and UpSC are integrated into MNAS-Unet, corresponding to downward and upward movements, respectively. A blue arrow indicates the progression of the feature map (input image). Additionally, the light purple arrow indicates a transformation operation within UpSC that is also subject to automatic search. Adapted from~\cite{weng2019unet}.}
\label{fig:NAS-Unet}
\end{figure*}

In this section, we describe the MNAS-Unet architecture as a Directed Acyclic Graph (DAG), where nodes represent feature maps and edges correspond to operations such as convolution or pooling. The architecture comprises two cell types, DownSC and UpSC, arranged symmetrically to form a U-shaped structure (Fig.~\ref{fig:NAS-Unet}). Each cell receives inputs from the previous two layers and supports operations like pooling, convolution and concatenation (Fig.~\ref{fig:an_example}). The network includes $L = 2L_1 = 8$ cells, evenly split between contraction and expansion paths, enabling accurate spatial recovery and dense probability map generation \cite{al2021cmm}.

The search space for MNAS-Unet consists of 6 Down POs (avg pool, max pool, down conv, down dep conv, down dil conv, down cweight), 4 Up POs (up cweight, up dep conv, up conv, up dil conv), and 6 Normal POs (identity, dep conv, dil conv, cweight, conv, shuffle conv). We follow the standard cell-based formulation with two input nodes and $M=4$ intermediate nodes per cell. Each intermediate node selects a small number of incoming edges and associated operations from the candidate set, resulting in a large yet structured discrete search space. This motivates the use of an efficient tree-based exploration strategy (MCTS) together with low-fidelity evaluation to reduce the practical search cost.

\begin{center}
\includegraphics[width=0.8\linewidth]{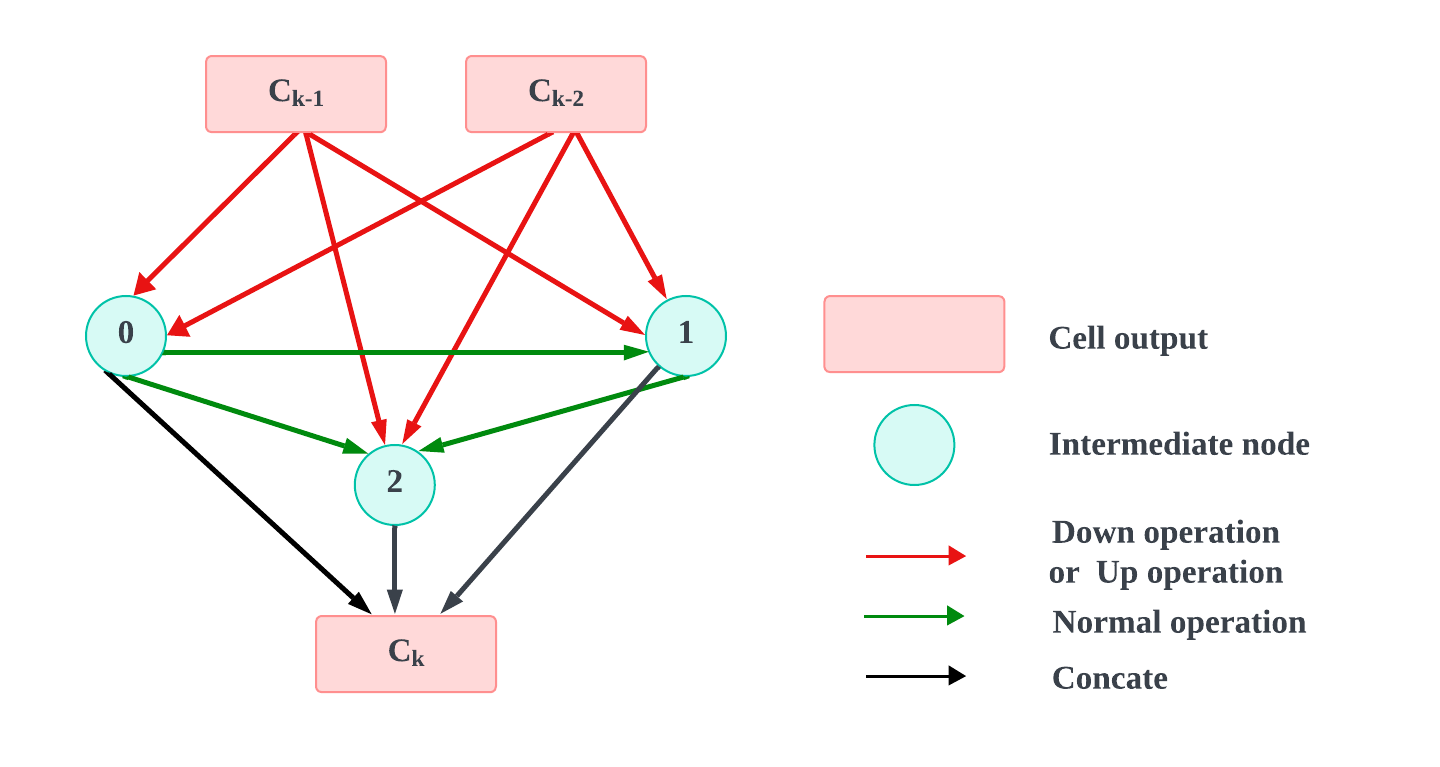}
\captionof{figure}{A depiction of the cellular architecture is provided. A red arrow indicates a reduction process, typically represented by max pooling. A green arrow marks a standard operation, which might be an identity or a convolution that maintains the dimensions of the feature map. Moreover, a black arrow highlights the concatenation operation. Adapted from~\cite{weng2019unet}.}
\label{fig:an_example}
\end{center}

\subsubsection{The Search Space of MNAS-Unet}
MNAS-Unet features an encoder-decoder architecture composed of customized units (cells) formed by interconnected nodes. The model defines three types of operations: Normal Operations for general feature extraction, Downward Operations for downsampling, and Upward Operations for upsampling, as shown in Table~\ref{tab:search_space}.

\begin{table}[tbp]
\centering
\caption{Search space for MNAS-Unet architecture search. Operations are selected based on genotype analysis from experimental results. The search space includes 6 Down POs, 4 Up POs, and 6 Normal POs.}
\label{tab:search_space}
\resizebox{\linewidth}{!}{%
\begin{tabular}{c|c|c|c}
\hline
\textbf{POs type} & \textbf{Down POs} & \textbf{Up POs} & \textbf{Normal POs} \\
\hline
1 & avg pool & up cweight & identity \\
\hline
2 & max pool & up dep conv & dep conv \\
\hline
3 & down conv & up conv & dil conv \\
\hline
4 & down dep conv & up dil conv & cweight \\
\hline
5 & down dil conv & -- & conv \\
\hline
6 & down cweight & -- & shuffle conv \\
\hline
\end{tabular}
}
\end{table}

\subsubsection{The Search Process of MNAS-Unet}
This section presents the application of the Monte Carlo Tree Search algorithm for neural architecture search within the MNAS-Unet framework. In this approach, MCTS evaluates the most promising actions at each state, while the MNAS-Unet architecture learns the performance characteristics of sampled architectures and generalizes this knowledge to unexplored architectural configurations \cite{wang2020neural}. This enables MCTS to perform multiple simulation rollouts by training the network only when evaluating previously unvisited nodes. MCTS allows MNAS-Unet to effectively balance exploration and exploitation, resulting in accelerated and more efficient discovery of optimal neural network architectures. This is accomplished through utilization of visitation statistics maintained at each node, which encompass: 1) $n(\sigma, \alpha)$, the visitation count for action $\alpha$ at state $\sigma$, and 2) $q(\sigma, \alpha)$, the expected reward for executing action $\alpha$ at state $\sigma$, representing the potential value of that trajectory in the search space. Fig.~\ref{fig:mnas_unet_process} demonstrates a typical search iteration within MNAS-Unet, comprising the phases of Selection, Expansion, MNAS-Unet architecture assisted Simulation, and Backpropagation, detailed as follows:

\textbf{Selection}: The selection phase traverses the search tree to identify the most promising path based on accumulated knowledge. Beginning from the root node, the traversal continues until reaching a terminal node. At each decision point, actions are selected using the Upper Confidence Bound 1 (UCB1) algorithm \cite{xu2023ucb}:
\begin{equation}
\pi_{\text{search}}(\sigma) = \arg \max_{\alpha \in \mathcal{A}} \left( \frac{q(\sigma, \alpha)}{n(\sigma, \alpha)} + 2c \sqrt{\frac{2 \log n(\sigma)}{n(\sigma, \alpha)}} \right)
\end{equation}
Here, $n(\sigma)$ represents the total visitation count for state $\sigma$, calculated as $n(\sigma) = \sum_{\alpha \in \mathcal{A}} n(\sigma, \alpha)$, and $c$ is a constant that balances exploration and exploitation. The first term, $q(\sigma, \alpha)/n(\sigma, \alpha)$, estimates the expected reward for an action, while the second term promotes exploration of less frequently visited nodes. When $n(\sigma, \alpha)$ is low, exploration is prioritized, while higher values of $n(\sigma, \alpha)$ favor exploitation. Initially, exploration is emphasized to gather sufficient information for making confident decisions in later stages \cite{el2023explore}. The exploration-exploitation balance is controlled by the constant $c$, typically set to 0.5 to moderate the growth of the search tree \cite{jerebic2021novel}.

\textbf{Expansion}: Upon reaching a leaf node, a new node is added to the tree. At this stage, both $q(\sigma, \alpha)$ and $n(\sigma, \alpha)$ are initialized to zero, and they will be updated during the subsequent simulation phase.

\textbf{MNAS-Unet Simulation Strategy}: The simulation phase evaluates the potential of an action by simulating future states from the expanded node. This is achieved by sampling offspring nodes and computing the $q(\sigma, \alpha)$ value based on their accuracies. Starting from the newly expanded node, random actions are selected until a terminal state is reached, and the resulting architecture is trained. The number of simulations directly affects the precision of the $q(\sigma, \alpha)$ estimate, with the formula for updating $q(\sigma, \alpha)$ given as:
\begin{equation}
q(\sigma, \alpha) \leftarrow \left( \text{Acc}\left( \text{sim}_0(\sigma') \right) + \frac{1}{k} \sum_{i=1}^{k} \text{Pred}\left( \text{sim}_i(\sigma') \right) \right) / 2
\end{equation}
Here, $\sigma' = \sigma + \alpha$ represents the new state after action $\alpha$, with $\text{sim}(\sigma')$ initiating a simulation from $\sigma'$. The term $\text{Acc}$ represents the training accuracy obtained from the first simulation, while $\text{Pred}$ represents the predicted accuracies in subsequent $k$ simulations by NAS-Unet. This approach refines the exploration process by prioritizing directions that are similar to previously successful architectures, thereby enhancing the exploitation aspect of the search.

\textbf{Backpropagation}: During backpropagation, the statistics $q(\sigma, \alpha)$ and $n(\sigma, \alpha)$ are updated along the path from the newly expanded node back to the root. Using the estimated reward $r$ for the new node, this information is iteratively propagated to its ancestor nodes:
\begin{equation}
\begin{aligned}
q(\sigma, \alpha) &\leftarrow q(\sigma, \alpha) + r, \\
n(\sigma, \alpha) &\leftarrow n(\sigma, \alpha) + 1, \\
\sigma &\leftarrow \text{parent}(\sigma), \quad \alpha \leftarrow \pi_{\text{search}}(\sigma)
\end{aligned}
\end{equation}
This process repeats until the root node is updated, ensuring that the cumulative experience gained from expanding new nodes is utilized to guide future exploration effectively \cite{wang2020neural}.

\begin{center}
\includegraphics[width=1.0\linewidth]{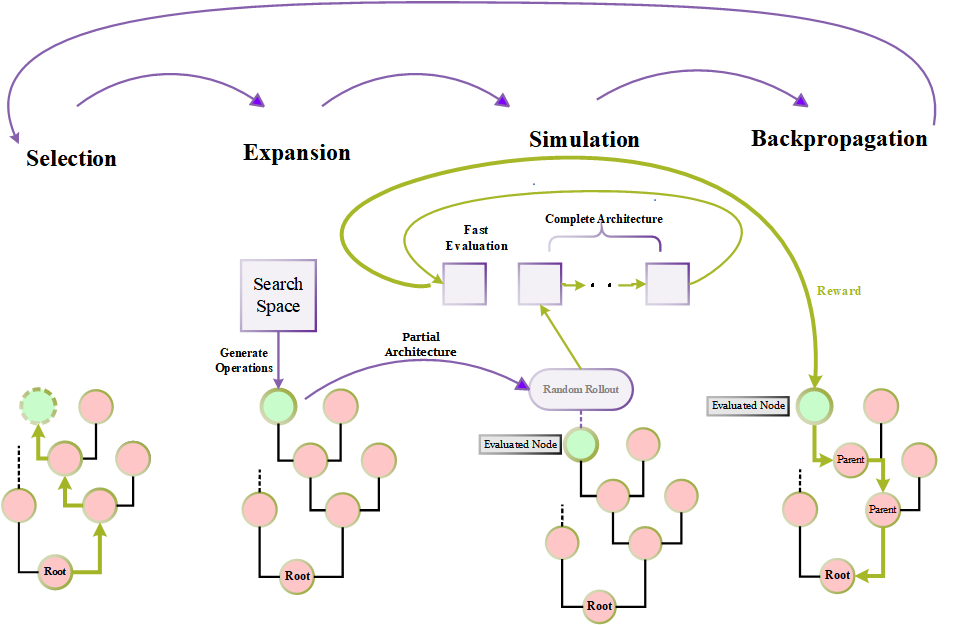}
\captionof{figure}{Overview of MNAS-Unet search procedures.}
\label{fig:mnas_unet_process}
\end{center}

\FloatBarrier
\section{Experimental Setup}
\label{sec4}

\subsection{Search Architecture Design}
MNAS-Unet uses $M=4$ intermediate nodes in both DownSC and UpSC, and the final network contains $L=2L_1=8$ cells arranged in a symmetric U-shaped backbone. We adopt the same cell-based representation as prior U-shaped NAS baselines, and use MCTS to efficiently explore the resulting discrete search space under a limited computational budget.

\subsection{Experimental Configurations}
To evaluate MNAS-Unet, we use the PASCAL VOC 2012 dataset and three medical imaging datasets: PROMISE12 \cite{zhang2023self}, CHAOS \cite{kavur2021chaos}, and Ultrasound Nerve Segmentation \cite{wang2023application}, which include MRI, CT, and ultrasound images. The model is optimized by minimizing the Dice Loss, derived from the negative Dice Similarity Coefficient (DSC), and performance is measured using DSC and Mean Intersection over Union (mIoU).

We compare MNAS-Unet with several baseline models, including U-Net, FC-Densenet, NAS-Unet, DC-Unet, DCSAU-Net, MNet, MedNeXt and ResTransUNet. All models were re-implemented in PyTorch \cite{paszke2019pytorch} to ensure consistency. The same data augmentation techniques were applied across all models, with additional preprocessing for noisy images \cite{zhou2023underwater,lu2018low}. Experiments were conducted on a system equipped with an NVIDIA RTX 4090 GPU, running Python 3.8 and PyTorch 1.11.0, with CUDA 11.3 support.

\subsection{Evaluation metrics}
We evaluate the performance of our model using several metrics, including Dice Similarity Coefficient (DSC), Validation Loss (Dice Loss), Pixel Accuracy (PA), and Intersection over Union (IoU), which are computed as shown in Eq.~\ref{eq:DSC}, \ref{eq:loss}, \ref{eq:PA}, and \ref{eq:IOU}.
\begin{equation}
DSC = \frac{2TP}{2TP + FP + FN}
\label{eq:DSC}
\end{equation}
\begin{equation}
DiceLoss = 1-\frac{2TP}{2TP + FP + FN}
\label{eq:loss}
\end{equation}
\begin{equation}
PA = \frac{TP + TN}{TP + TN + FP + FN}
\label{eq:PA}
\end{equation}
\begin{equation}
IOU = \frac{TP}{TP + FP + FN}
\label{eq:IOU}
\end{equation}
where $TP$ refers to True Positive, $TN$ refers to True Negative, $FP$ refers to False Positive, and $FN$ refers to False Negative.

\subsection{Dataset Overview}
This section introduces four widely-used datasets in the field of computer vision and medical image analysis: PASCAL VOC 2012, which supports a variety of tasks including object detection, classification, and semantic segmentation; PROMISE12, focused on prostate MRI segmentation; the 2016 Kaggle Ultrasound Nerve Segmentation dataset for Brachial Plexus nerve segmentation; and the Chaos Challenge cross-modality dataset, which includes tasks such as abdominal organ segmentation using both CT and MRI images.

The PASCAL VOC 2012 dataset is a cornerstone resource for various computer vision tasks such as image classification, object detection, and semantic segmentation. It consists of 1,464 training images and 1,449 validation images, categorized into 20 classes, including humans, animals, vehicles, and indoor objects. The images vary in size from $100 \times 100$ to $500 \times 500$ pixels, with the majority being approximately $375 \times 500$ pixels. Each pixel in these images is annotated with one of the 20 object categories or the background, providing a high-quality foundation for model training.

The PROMISE12 dataset is a specialized resource for prostate MRI segmentation, originally designed for the 2012 MICCAI challenge. It contains 50 training cases, each with transverse T2-weighted MRI slices of the prostate, totaling around 1,250 labeled images. These images are binary, with voxel values of either 0 or 1, and are resized to $256 \times 256$ pixels. To enhance image contrast, the dataset has been processed with Contrast Limited Adaptive Histogram Equalization (CLAHE).

The 2016 Kaggle Ultrasound Nerve Segmentation challenge focused on the segmentation of the Brachial Plexus (BP) nerves in ultrasound images. This dataset includes 5,635 training images and 5,508 test images, with the test set divided into 20\% for public ranking and 80\% for final evaluation. Notably, around 60\% of the training images do not contain the BP region. The images are standardized to a resolution of $580 \times 420$ pixels, though the dataset is known for inconsistencies, particularly with images that have conflicting labels despite appearing nearly identical. For instance, one image may have a non-empty mask, while its duplicate might have an empty mask, as illustrated in Fig.~\ref{fig:contradictory_images}.

Lastly, the Chaos Challenge, which took place at the 2019 IEEE International Symposium on Biomedical Imaging (ISBI) in Venice, Italy, featured two distinct databases: Abdominal CT and MRI. This paper focuses on the cross-modality dataset, which presents the challenging task of liver segmentation from CT images and abdominal organ segmentation from MRI images. Both training and validation sets consist of images from these two modalities, contributing to the dataset's complexity.

\begin{center}
\begin{minipage}[t]{0.48\linewidth}
\centering
\includegraphics[width=\textwidth]{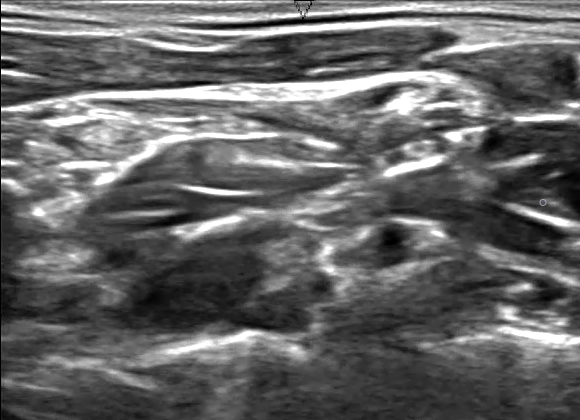}\\[0.3em]
\textbf{(a)}
\end{minipage}
\hfill
\begin{minipage}[t]{0.48\linewidth}
\centering
\includegraphics[width=\textwidth]{images/nerve28.png}\\[0.3em]
\textbf{(b)}
\end{minipage}

\vspace{0.8em}

\begin{minipage}[t]{0.48\linewidth}
\centering
\includegraphics[width=\textwidth]{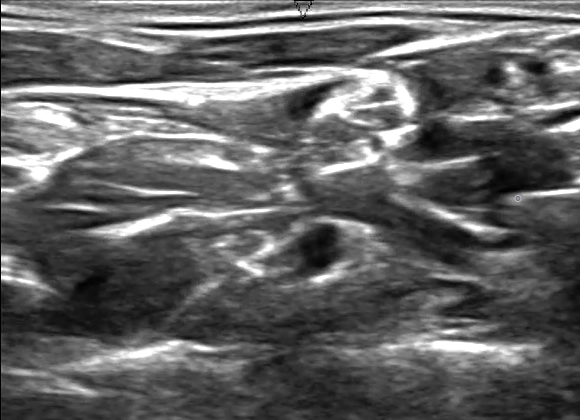}\\[0.3em]
\textbf{(c)}
\end{minipage}
\hfill
\begin{minipage}[t]{0.48\linewidth}
\centering
\includegraphics[width=\textwidth]{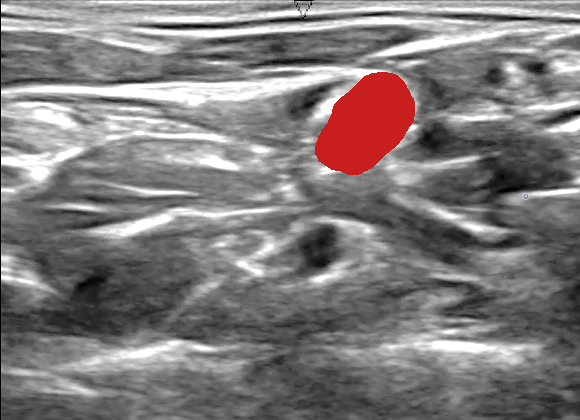}\\[0.3em]
\textbf{(d)}
\end{minipage}

\vspace{0.5em}
\captionof{figure}{Subfigures (a)(b) show the same image without a segmentation mask; (c)(d) show the same case with the mask overlaid in (d). Such inconsistencies are common in human-labeled data.}
\label{fig:contradictory_images}
\end{center}

\subsection{Experimental Results}
This paper proposed an efficient cell search strategy \cite{yang2020cars} for medical image segmentation and built a lightweight model with fewer parameters based on this strategy \cite{zhou2020review}. We investigated proxy cell architectures using the more intricate PASCAL VOC 2012 dataset \cite{tong2023rethinking} and then adapted these architectures for medical image datasets. We evaluated the efficiency and performance of MNAS-Unet and NAS-Unet in neural network architecture searches.

In the PASCAL VOC 2012 dataset, we split the search stage into 1,464 images for training and 1,449 for validation. We performed architecture search by comparing the ProxylessNAS strategy combined with Monte Carlo Tree Search (MCTS) to the traditional ProxylessNAS, using a batch size of 16 and setting the training period to 300 epochs. The network weights $w$ were trained with the AdaBound optimizer, initialized with a learning rate of 0.0005 and weight decay of 0.0005, alongside a cosine learning rate schedule. For architectural parameters $\alpha$, we utilized the Adam optimizer with an initial learning rate of 0.0001 and weight decay of 0.005, optimizing $\alpha$ from the start.

The early termination of MNAS-Unet at 139 epochs due to the patience limit indicates improved search efficiency. Under the same search setting, NAS-Unet runs for the full 300 epochs, whereas MNAS-Unet converges earlier (Fig.~\ref{fig:mnas_unet} vs.\ Fig.~\ref{fig:nas_unet}). In terms of training-epoch budget (and corresponding GPU compute), this reduces the search cost by $1 - 139/300 \approx 54\%$.

\begin{center}
\begin{minipage}[t]{0.48\linewidth}
\centering
\includegraphics[width=\linewidth]{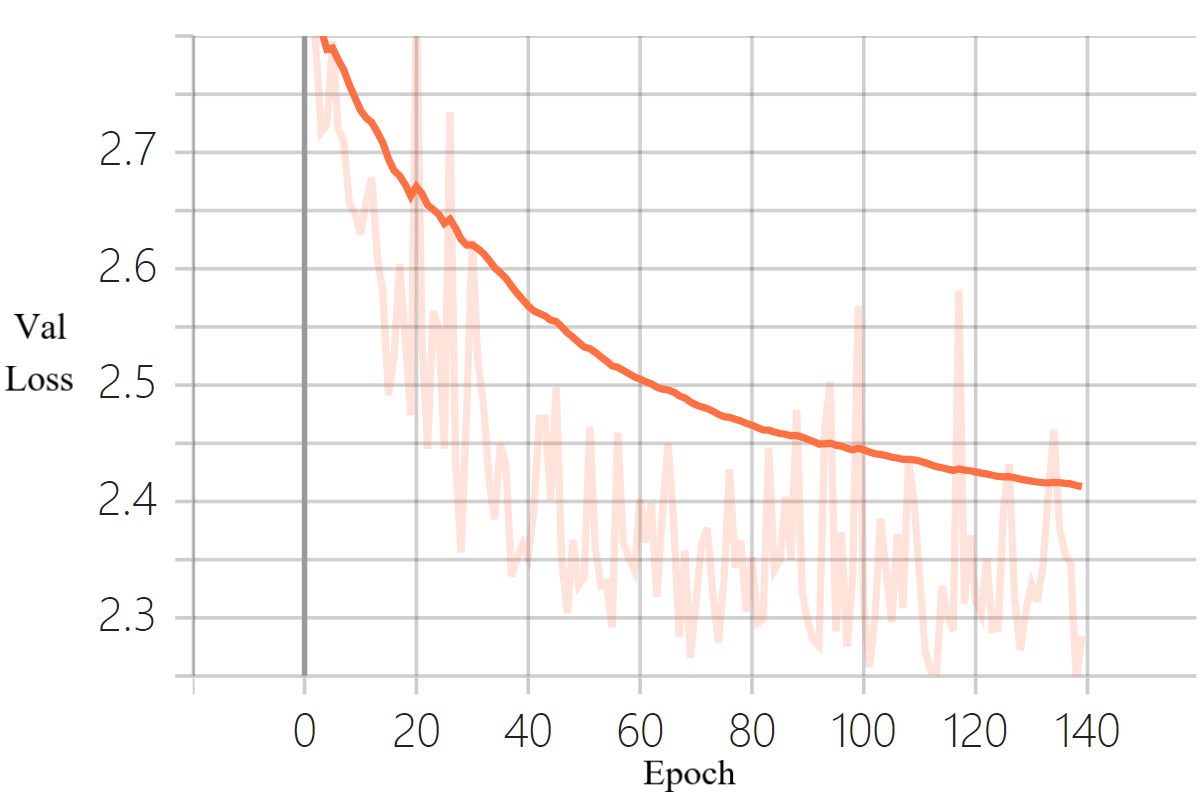}\\[0.3em]
\textbf{(a)}
\end{minipage}
\hfill
\begin{minipage}[t]{0.48\linewidth}
\centering
\includegraphics[width=\linewidth]{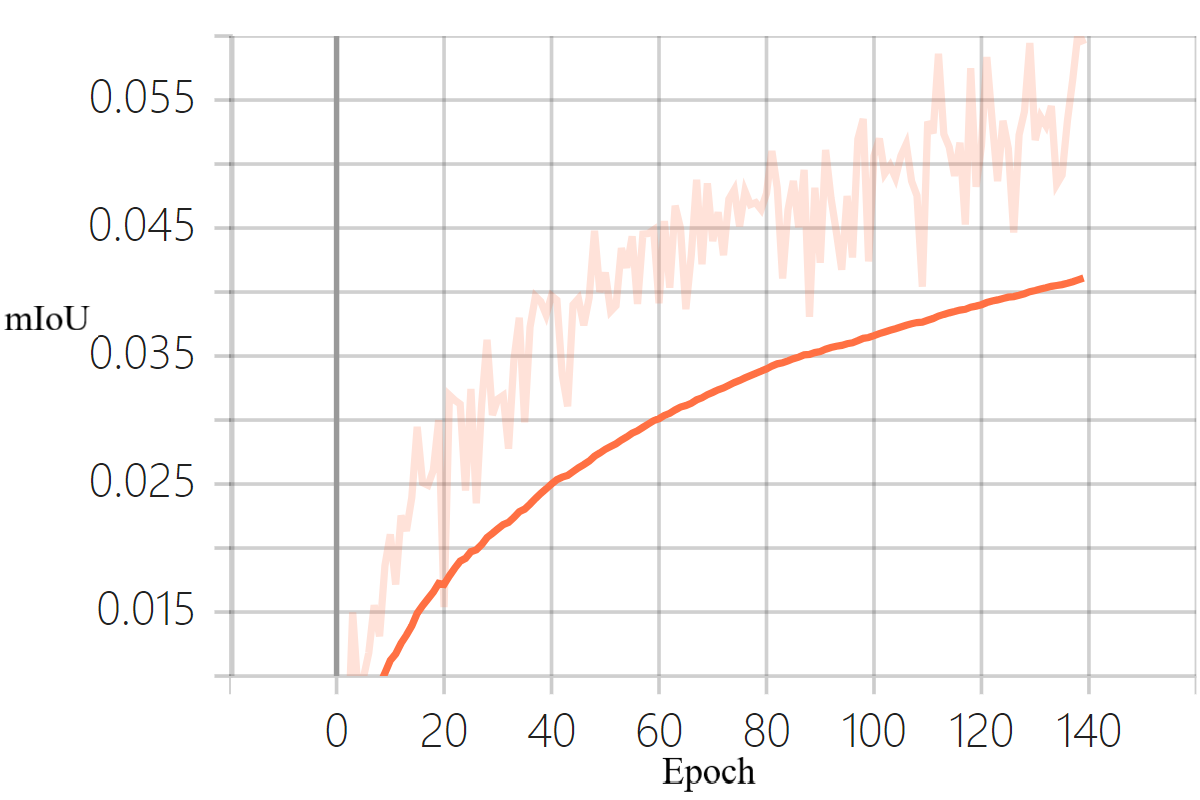}\\[0.3em]
\textbf{(b)}
\end{minipage}

\vspace{0.5em}
\captionof{figure}{The validation loss and mIoU of candidate best architecture network of MNAS-Unet on PASCAL VOC 2012 dataset.}
\label{fig:mnas_unet}
\end{center}

\begin{center}
\begin{minipage}[t]{0.48\linewidth}
\centering
\includegraphics[width=\linewidth]{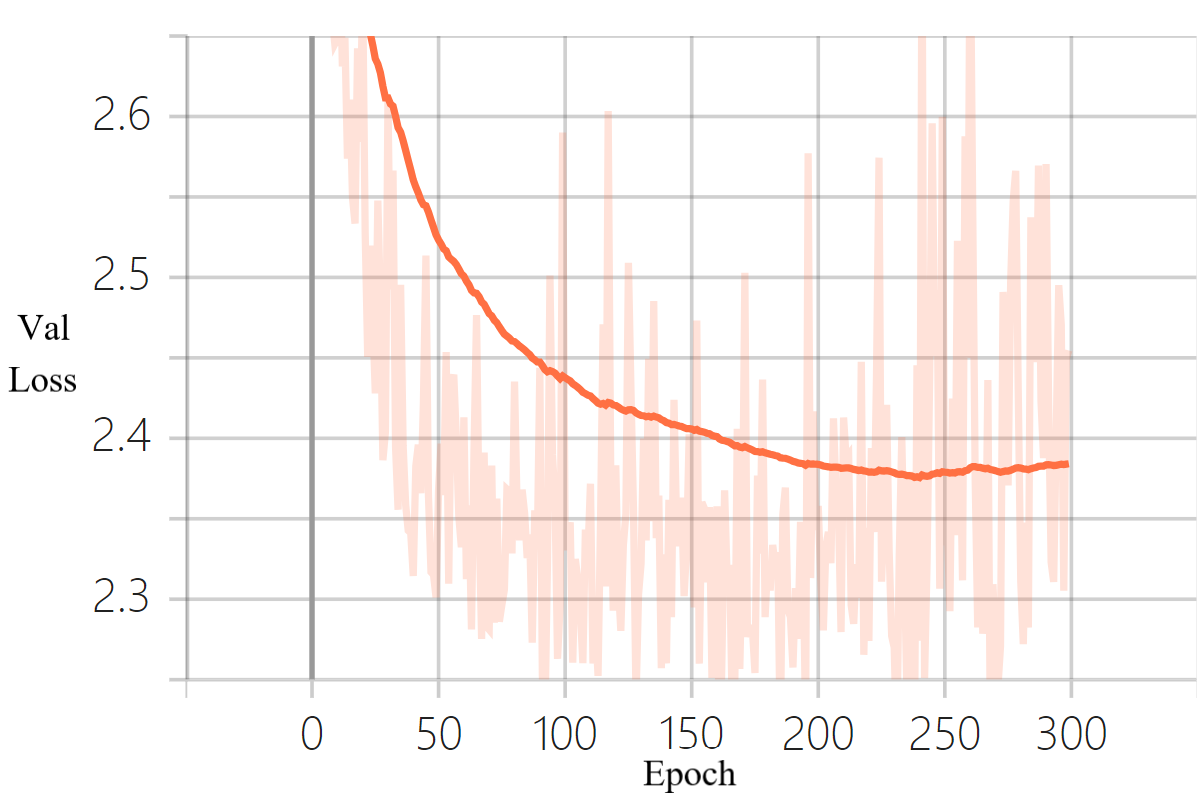}\\[0.3em]
\textbf{(a)}
\end{minipage}
\hfill
\begin{minipage}[t]{0.48\linewidth}
\centering
\includegraphics[width=\linewidth]{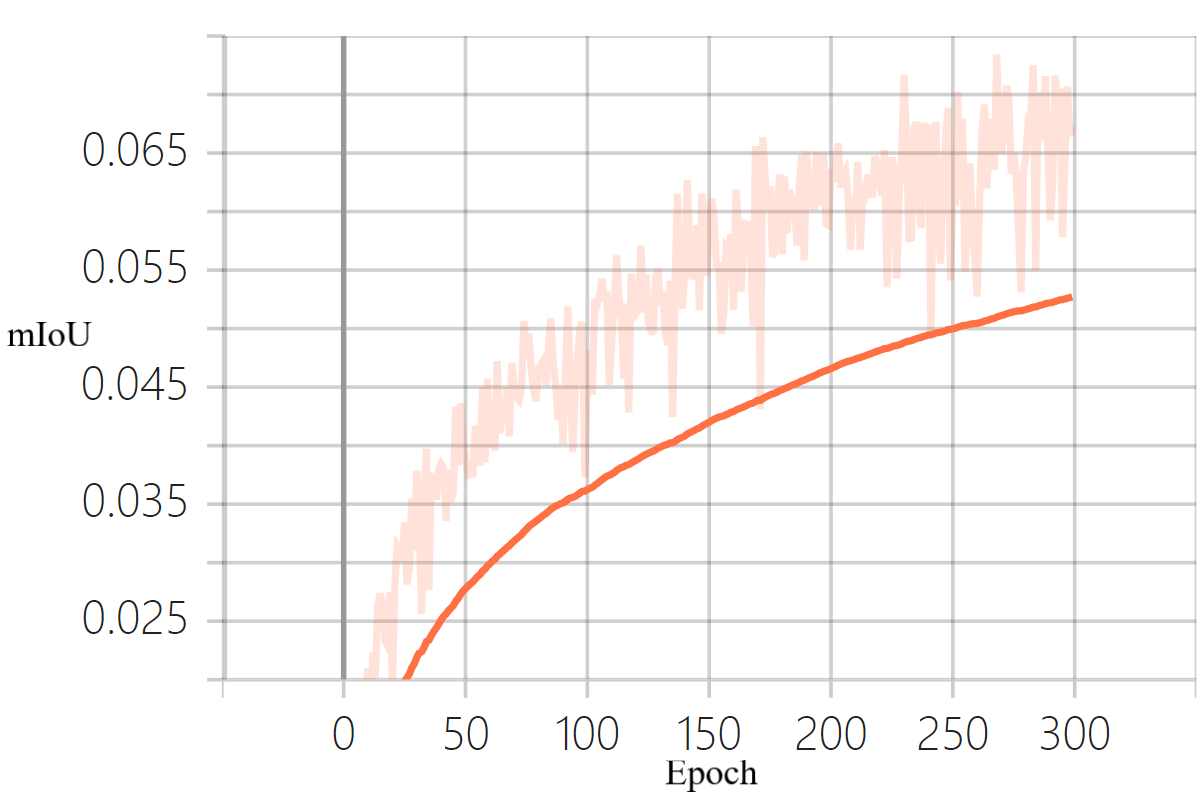}\\[0.3em]
\textbf{(b)}
\end{minipage}

\vspace{0.5em}
\captionof{figure}{The validation loss and mIoU of candidate best architecture network of NAS-Unet on PASCAL VOC 2012 dataset.}
\label{fig:nas_unet}
\end{center}

We compared the performance of MNAS-Unet, NAS-Unet, and several state-of-the-art models across different medical image datasets, including PROMISE12, Ultrasound Nerve Segmentation, and Chaos. The experimental results are summarized in Table~\ref{tab:performance_comparison}. Please note that GM refers to the cost of GPU memory. GM is measured as the peak GPU memory allocated during training (reported by \texttt{torch.cuda.max\_memory\_allocated}) under the same batch size and input resolution.

\begin{table*}[tbp]
\centering
\caption{Performance comparison across datasets. MNAS-Unet achieves the best overall segmentation performance while maintaining lower memory consumption. Training time refers to single-run wall-clock time measured under identical batch size and input resolution.}
\label{tab:performance_comparison}
\resizebox{\textwidth}{!}{%
\begin{tabular}{|c|c|c|c|c|c|c|c|c|c|c|c|c|}
\hline
\multirow{2}{*}{\textbf{Model}}
& \multicolumn{4}{c|}{\textbf{PROMISE12}}
& \multicolumn{4}{c|}{\textbf{Ultrasound}}
& \multicolumn{4}{c|}{\textbf{CHAOS}} \\
\cline{2-13}
& mIoU & DSC & Time & GM
& mIoU & DSC & Time & GM
& mIoU & DSC & Time & GM \\
\hline
U-Net
& 0.515 & 0.680 & 46min & 6.5GB
& 0.662 & 0.797 & 6h05m & 7.0GB
& 0.886 & 0.939 & 13h08m & 8.0GB \\
\hline
FC-DenseNet
& 0.528 & 0.691 & 53min & 7.2GB
& 0.673 & 0.805 & 6h28m & 7.4GB
& 0.902 & 0.948 & 13h22m & 8.0GB \\
\hline
DC-UNet
& 0.533 & 0.695 & 49min & 6.5GB
& 0.676 & 0.808 & 6h02m & 6.9GB
& 0.898 & 0.946 & 13h05m & 7.5GB \\
\hline
DCSAU-Net
& 0.539 & 0.701 & 51min & 6.8GB
& 0.681 & 0.811 & 6h14m & 7.0GB
& 0.905 & 0.952 & 13h12m & 7.7GB \\
\hline
MNet
& 0.534 & 0.696 & 48min & 6.2GB
& 0.684 & 0.813 & 5h47m & 6.6GB
& 0.907 & 0.953 & 12h34m & 7.3GB \\
\hline
MedNeXt
& 0.547 & 0.708 & 44min & 6.6GB
& 0.693 & 0.820 & 5h52m & 6.7GB
& 0.912 & 0.955 & 12h37m & 7.4GB \\
\hline
ResTransUNet
& 0.556 & 0.715 & 47min & 6.4GB
& 0.701 & 0.824 & 6h08m & 7.1GB
& 0.918 & 0.958 & 13h03m & 7.6GB \\
\hline
NAS-Unet
& 0.561 & 0.719 & 39min & 6.5GB
& 0.714 & 0.833 & 5h20m & 6.8GB
& 0.927 & 0.962 & 12h07m & 7.5GB \\
\hline
\textbf{MNAS-Unet}
& \textbf{0.573} & \textbf{0.728} & \textbf{36min} & \textbf{5.8GB}
& \textbf{0.727} & \textbf{0.842} & \textbf{4h41m} & \textbf{6.3GB}
& \textbf{0.933} & \textbf{0.966} & \textbf{11h52m} & \textbf{7.0GB} \\
\hline
\end{tabular}}
\end{table*}

\FloatBarrier
\section{Discussion and Conclusion}
\label{sec5}

In this paper, Monte Carlo Tree Search (MCTS) is integrated into Neural Architecture Search (NAS) for medical image segmentation, resulting in the proposed MNAS-Unet framework. By combining MCTS-guided exploration with a tailored cell search space on a U-shaped backbone, MNAS-Unet improves search efficiency and achieves strong segmentation performance across multiple modalities. To cope with the complexity and diversity of medical images, we design three basic operation sets with special emphasis on stacked DownSC and UpSC cells, aiming to enhance the adaptability of the model to different medical images such as CT and MRI \cite{chen2020unsupervised}. Compared with the traditional ProxylessNAS method, incorporating MCTS significantly improves the architecture search efficiency and optimizes the gradient descent process, thus accelerating the design of the network architecture and enhancing the segmentation accuracy.

Experimental results demonstrate the superiority of MNAS-Unet across multiple datasets, including PROMISE12, Chaos, and ultrasound segmentation. MNAS-Unet outperforms NAS-Unet and other state-of-the-art models in key metrics, showcasing the effectiveness of MCTS in improving segmentation accuracy, particularly for medical images with complex structures and textures.

Our contributions are the integration of MCTS with NAS for medical image segmentation, the design of a tailored search space (6 Down, 4 Up, and 6 Normal POs), and the demonstration of improved search efficiency and segmentation accuracy across multiple datasets, with applications in emergency diagnostics, portable ultrasound devices, and large-scale screening programs.

Future work will focus on enhancing the interpretability of the MNAS-Unet model, particularly in explaining decision-making processes and identifying key features relevant to clinical diagnoses, to improve its trustworthiness and usability in real-world medical environments.

\bio{}
Liping Meng received the B.Eng. degree in Computer Science and Technology from Yan'an University, China, in 2022, and the M.Sc. degree in Applied Mathematics from Saint Petersburg State University, Russia, in 2024. She is currently an Assistant Lecturer with the School of Science and Engineering at Xi'an Kedagaoxin University, China. Her research interests include medical image analysis and processing, efficient deep learning, and lightweight large-scale neural network models.
\endbio

\bio{}
Fan Nie is a university lecturer with a graduate degree from the University of Hong Kong and research training at the Institute of Psychology, Chinese Academy of Sciences. Her work sits at the intersection of educational research and computational methods. With a strong interest in how technology mediates learning, she focuses on academic writing that explores the application of computer-based tools in understanding and enhancing educational processes. She is particularly skilled at bridging complex ideas across disciplines to foster clearer scholarly communication.
\endbio

\bio{}
Yunyun Zhang is an Associate Professor who received the M.A. degree from Northwestern Polytechnical University, China. She was a government-sponsored visiting scholar in the United States and a recipient of the Fulbright Scholarship. Her research interests include English studies, translation, artificial intelligence–assisted language engineering, and intelligent English education, with particular emphasis on interdisciplinary integration of language and artificial intelligence. She has extensive experience in higher education teaching and research and has published widely in related fields. She is currently a faculty member at Xi'an Kedagaoxin University.
\endbio

\bio{}
Chao Han is a Senior Engineer and currently serves as the Director of the Computer Science Teaching and Research Office at Xi'an Kedagaoxin University. He has published several academic papers and holds 17 patents. He has supervised students to achieve multiple awards in national and provincial academic competitions and has been recognized as an outstanding instructor at both levels. His research interests include artificial intelligence, embedded systems, microcontroller-based systems, intelligent hardware design, and applied machine learning.
\endbio

\end{document}